# Persian Signature Verification using Fully Convolutional Networks

**Mohammad Rezaei[1,*] - Nader Naderi[2]**

[1] Department of computer engineering, K.N.Toosi University of Technology, Tehran, Iran

*corresponding author: mohmd.rezaei92@email.kntu.ac.ir, telephone number: 00989171704882

[2] Department of engineering, Islamic Azad University Arak Branch, Iran

Email address: ndr.naderi@gmail.com

**ABSTRACT**

*Fully convolutional networks (FCNs) have been recently used for feature extraction and classification in image and speech recognition, where their inputs have been raw signal or other complicated features. Persian signature verification is done using conventional convolutional neural networks (CNNs). In this paper, we propose to use FCN for learning a robust feature extraction from the raw signature images. FCN can be considered as a variant of CNN where its fully connected layers are replaced with a global pooling layer. In the proposed manner, FCN inputs are raw signature images and convolution filter size is fixed. Recognition accuracy on UTSig database, shows that FCN with a global average pooling outperforms CNN.*

## 1. INTRODUCTION

Signature is a biometric characteristic that is used in person authentication [1]. Automatic signature verification is an interesting area of research, since biometric authentication is more trustable alternative to password based security systems. Biometric authentication is widely being used as it is relatively hard to be forgotten, stolen, or guessed [2]. Numerous biometric features have been studied and proved useful, including biological characteristics such as fingerprint, face, iris, and retina pattern or behavioral traits such as signature and speech [2-4].

Signature authentication can be considered as a low cost biometric system where awareness and uniqueness of person is necessary [2]. There are two main research fields in this area: signature recognition (or identification) and signature verification. The signature recognition involves on identifying the author of a signature when a signature database is searched to find the identity of a given signer. While signature verification defines the process of testing a signature to decide whether a particular signature truly belongs to a person or not. In this case, the output is either accepting the signature as valid or rejecting it as a forgery one [2,5].

Signature verification systems are classified either online or offline depending on the data acquisition method and involved application. Usually, online signature verification systems present a better performance than the offline signatures verification systems. In the online approach, the signature is captured using a special input device and the system uses the signature as well as the dynamic information obtained during the signing process (pen's position, inclination). However, in online signature verification system, the presence of the signer is required at both times of obtaining the reference signature and the verification process which is not welcomed by many applications. Consequently, offline verification methods have more practical application areas than the former. The offline approach only uses the digitalized image of a signature extracted from a document called static information. Therefore, it does not require any special processing devices. On the other hand, preprocessing is more difficult and time consuming in offline systems due to unavailability of the dynamic information [2,5,6].

Various techniques have already been applied in signature verification such as fuzzy logic [7], geometric features [8, 9], global characteristics [10], genetic algorithms [11], neural networks [12-14], hidden Markov models [15], discrete wavelet transform (DWT) and image fusion [16], dynamic time warping-based segmentation and Multivariate autoregressive model [17], convolutional neural networks (CNNs) and its variants [2,18], support vector machine (SVM) classifier and fixed-point arithmetic [19]. Clearly, most of researches in offline signature verification are involved in feature extraction.

Persian signatures are different from other nation signatures, since people usually do not use text in it and they draw a shape as their signature. Even if they use text in their signature, mostly it is in Persian which is so hard to distinguish using machine learning approaches. Persian signature verification has been done using DWT and image fusion, SVM classifier and fixed-point arithmetic, dynamic time warping-based segmentation and Multivariate autoregressive model and convolutional neural networks [2,16,17,19,20].

Convolutional neural networks can be considered as a kind of the standard neural networks consisting of alternating convolution and pooling layers [21]. CNNs can be compared to fully connected deep neural networks (DNNs) due to its two main properties. First, CNNs have much fewer connections and parameters due to the local-connectivity and shared-filter architecture in the convolution layers. Second, pooling at a local region results in generalization using a form of translation invariance [21].

There are many researches on feature extraction using CNNs for speech recognition, image processing and signature verification [2,21,22]. Fully convolutional networks (FCNs), without fully connected layers, have been successfully applied to other fields like speech and image processing [23-25].

In this paper, we propose to use fully convolutional neural network for extracting robust features from the signature image.

Remainder of this paper is organized as follows. Section 2 introduces CNN briefly. Section 3 includes our proposed method. Section 4 contains our experimental results. Finally, we give our conclusion in Section 5.

## 2. CONVOLUTIONAL NEURAL NETWORK

### 2.1 Architecture

The standard Convolutional Neural Network consists of three essential parts: first two parts are alternating convolution and pooling layers, and the third one is fully connected layers on top of the last pooling layer.

### 2.2 Convolution Layer

In order to have an obvious observation about the convolution and pooling layers operations, a simple network architecture, borrowed from [26], have been shown in Figure. 1. A convolution layer is assumed to be fed with a $t \times f$ signal where $t$ and $f$ are the input image dimensions [27]. The convolution layer is composed of $n$ neurons, which each neuron can be considered as a filter [21,25-27]. Each filter (kernel), numbered with $j$, has an $m \times r$ (called filter size) matrix of parameters replicated across the entire input space. This matrix is convolved with the input ($I$) and eventually produces a feature map of size $(t-m+1) \times (f-r+1)$, where $m<=t$ and $r<=f$, which can be shown as:

$$o_{(t-m+1) \times (f-r+1)}^{(l,j)} = k_{m \times r}^{(l,j)} * I_{t \times f}^l + b_{1 \times 1}^{(l,j)} ,  \qquad (1)$$

Where $o^{(l,j)}$ denotes $j$-th output of $l$-th layer, $k^{(l,j)}$ denotes the kernel weights of $j$-th neuron in $l$-th layer, $b^{(l,j)}$ is the bias of $j$-th neuron in $l$-th layer, and * indicates convolution operator [21]. As a result of convolution operation, kernel weights are shared among whole input features for each neuron [21,25-27].

This weight sharing reduces the number of network learned weights compared to the standard fully connected neural networks. On the other hand, it models local correlations in the input signal. It can also improve model robustness and reduce overfitting as each weight is learned from multiple frequency bands and time spaces in the input instead of just from one single location [21,25-27].

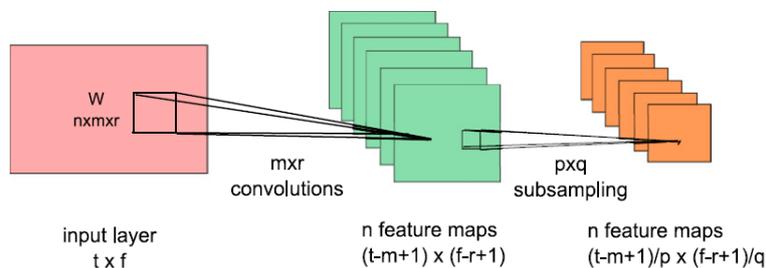

**Fig. 1.** *The operation of convolution and pooling layers* [27].

### 2.3 Pooling layer

After the convolution layer, a pooling layer is added to compute a lower resolution representation of the convolution layer feature maps through sub-sampling from different positions within a specified window. Several pooling strategies have been proposed, the best known pooling strategies are: average pooling and max pooling [21,26].

Max pooling is the most popular pooling strategy, which outputs the maximum value within each specified window [2,21,25-27]. Also, if the maximization function is replaced with an average function, the pooling strategy turns to average pooling [26]. Max pooling has been shown to give improvements in performance in image processing tasks compared to the average pooling [21,25].

Max pooling aims to handle variability and slight shifts that are common in image signal due to additive and channel noise [27]. As a result, it outputs a lower resolution of convolution feature maps and causes generalization [21].

Pooling can be used with or without overlap. Overlapped pooling has not produced a clear performance gain. So, using a pooling size of p × q (without overlap), there will be n neurons containing $\frac{t-m+1}{p} \times \frac{f-r+1}{q}$ feature maps [21,27].

Often, the convolution or pooling layer has been followed by a non-linear activation function such as Sigmoid or ReLU [2,21,25-27].

## 2.4 Fully Connected Layers

One or more fully connected hidden layers have been added to the top of the final pooling layer in order to combine the features across all frequency bands before feeding to the output layer [25].

## 2.5 Fully Convolutional Networks

Standard deep convolutional neural networks usually use fully connected layers of high dimensions (at least 512) on the top of the final pooling layer, leading to a very high number of parameters. It has been shown that most of the learning occurs in the convolutional layers, so convolutional neural networks without any fully connected layer could be effective. By removing fully connected layers, the CNN is forced to learn good representation in the convolution layers, which potentially leading to better generalization [21,23,24].

## 3. PROPOSED METHOD

As mentioned earlier, CNN has been used for Persian signature verification, where its input has been raw image of signature [2]. In this paper, we use fully convolutional network for extracting the features and classification. In this manner, a signature image is fed into the fully convolutional network where we expect that its output represents the decision on real signature or a forgery one. The common parameters and layers are considered like CNN structure in [2]. The proposed FCN is shown in Figure 2.

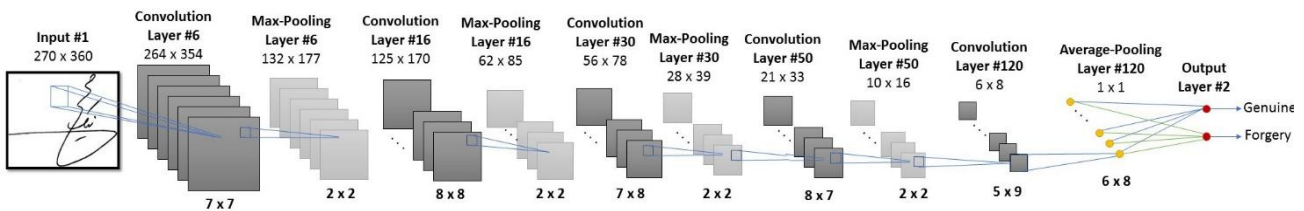

**Fig. 2.** *Block diagram of proposed FNN for signature verification.*

## 3.1 Pooling type

For regular pooling layers (on top of each convolution layer), like [2] max pooling is used. However, in order to preserve information in the last convolution feature maps, average pooling is used for the last pooling layer which is rather than fully connected layers.

## 3.2 Filter and Pooling sizes

Equivalent filter size for the convolution layers and pooling size for pooling layers on top of each convolution layer is selected based on [2]. Instead of fully connected layers, a single global average pooling layer, which reduces each output feature map from previous layer to one value, is added on top of last convolution layer and its outputs are fed into the classification layer. Consequently, the number of learning features are reduced and full advantages of convolution layers are consumed.

## 4 EXPERIMENTS AND RESULTS

### 4.1 Experimental setup

Experiments have been performed on UTSig [19] database. Training and testing data are selected like setup 1 in [19]. Since CNN needs a large amount of data, we have changed the number of selected images. In this way, training data is achieved by separating 25 randomly selected genuine samples of each person and 25 randomly selected simple forgery samples of each.

Testing data is created using 2 remaining genuine samples, along with remaining simple, skilled and opposite-hand forgeries.

All signature images are resized to $270 \times 360$. No further preprocessing is done since CNN is shown to be a good feature extractor. We have used stochastic gradient descent and mini-batch size of 128 for CNN and FCN training. The learning rate is equal to 0.001. The number of epochs have been selected equal to 100 which have been shown the best performance in our experiments. More epochs caused over fitting.

### 4.2 Results of CNN and FCN

Persian signature verification using baseline CNN is done on different dataset which is not in access and is not a sufficient amount to train and test DNN. As a result, a CNN exactly like [2] is trained and tested on UTSig and is compared with our proposed FCN and the results have been compared in Table 1.

For decision-based criteria, we calculate false acceptance rate (FAR) (i.e. the percent of forged samples that are incorrectly accepted), false rejection rate (FRR) (i.e. the percent of genuine samples that are incorrectly rejected), and accuracy (i.e. the percent of forged and genuine samples that are correctly labeled). As can be seen from Table 1, FCN has a better performance comparing with CNN.

### 4.3 Comparison between FCN and SVM

In order to evaluate overall systems, we have compared our results with the results of SVM from [19] in Table 2. As can be seen from Table 2 SVM performed better than FCN in FAR criteria even though in FRR criteria FCN performed so much better.

**Table 1.** *Results of accuracy, FAR and FRR for CNN and FCN.*

| Model | Accuracy | FAR | FRR |
|-------|----------|-------|-------|
| CNN | 65.06 | 37.83 | 29.69 |
| FCN | 76.71 | 27.47 | 15.72 |

**Table 2.** *Results of FAR and FRR for FCN and SVM.*

| Model | FAR | FRR |
|-------|-------|-------|
| FCN | 27.47 | 15.72 |
| SVM | 21.29 | 39.27 |

## 5 CONCLUSION

In this paper, we proposed to use FCN for feature extraction and classification from signature images. Although, CNN has been used for this purpose, fully connected layers in CNN are believed to prevent convolution layers from learning good representation. Therefore, we have used FCN where CNN fully connected layer is replaced with global average pooling. The experimental results show that FCN using a global average pooling outperforms CNN and is comparable with SVM method for extracting robust features from offline signature images and also classification.

### REFERENCES

[1] Gabe, A., Sheffer, B., and Bryant, M. (2016). Offline Signature Verification with Convolutional Neural Networks.

[2] Khalajzadeh, H., M. Mansouri, and M. Teshnehlab. (2012), Persian signature verification using convolutional neural networks. *International Journal of Engineering Research and Technology 1*.

[3] Ververidis, D., Kotropoulos, C. (2006), Emotional speech recognition: Resources, features, and methods. *In Speech Communication*, Volume 48, Issue 9, Pages 1162-1181, ISSN 0167-6393, https://doi.org/10.1016/j.specom.2006.04.003.

[4] Sandeep, D., and Mahesh, M. K. (2014), Handwriting Analysis of Human Behavior Based on Neural Network. *Int. J. Adv. Res. Comput. Sci. Softw. Eng 4.9*.

[5] Fahmy, M.M. (2010), Online handwritten signature verification system based on DWT features extraction and neural network classification. *In Ain Shams Engineering Journal*, Volume 1, Issue 1, Pages 59-70, ISSN 2090-4479, https://doi.org/10.1016/j.asej.2010.09.007.

[6] Hafemann, L.G., Sabourin, R., Oliveira, L.S. (2016), Writer-independent Feature Learning for Offline Signature Verification using Deep Convolutional Neural Networks. *The International Joint Conference on Neural Networks (IJCNN)*, doi. 10.1109/IJCNN.2016.7727521.

[7] Ismail, M.A., Gad, S. (2000), Off-line Arabic signature recognition and verification. *Pattern Recognition 33*, 1727–1740.

[8] Fang, B., Wang, Y.Y., Leung, C.H., Tang, Y.Y., Kwok, P.C.K., Tse, K.W., Wong, Y.K., (1999), A smoothness index based approach for off-line signature verification. *Proceedings of ICDAR"99*, 785–787.

[9] Hobby, J.D. (2000), Using shape and layout information to find signatures, text, and graphics. *Computer Vision and Image Understanding*, 80, 88– 110.

[10] Ramesh, V.E., Murty, M.N. (1999), Off-line signature verification using genetically optimized weighted features, *Pattern Recognition 32*, 217–233.

[11] Scholkopf, B., Sung, K., Burges, C., Girosi, F., Niyogi, P., Poggio, T., Vapnik, V. (1996), Comparing support vector machines with Gaussian kernels to radial basis function classifiers. *AI Memo* No. 1599, MIT.

[12] Bajaj, R., Chaudhury, S. (1997), Signature verification using multiple neural classifiers, *Pattern Recognition*, 30, 1–7.

[13] Baltzakis, H., Papamarkos, N. (2001), A new signature verification technique based on a two-stage neural network classifier, *Engineering Applications of Artificial Intelligence 14* 95–103.

[14] Velez, J.F., Sanchez, A., Moreno, A.B. (2003), Robust off-line signature verification using compression networks and positional cuttings, *Proceedings of IEEE International Conference on Neural Networks for Signal Processing* (NNSP „03). 627–636.

[15] Camino, J.L., Travieso, M.C., Morales, C.R., Ferrer, M.A. (1999), Signature classification by hidden Markov model, *Proceedings of IEEE International Carnahan Conference on Security Technology*. 481–484.

[16] Ghandali, S. and Moghaddam, M. E. (2008), A Method for Off-line Persian Signature Identification and Verification Using DWT and Image Fusion. *IEEE International Symposium on Signal Processing and Information Technology*, Sarajevo, pp. 315-319, doi: 10.1109/ISSPIT.2008.4775712.

[17] Zoghi, M. and Abolghasemi, V. (2009), Persian signature verification using improved Dynamic Time Warping-based segmentation and Multivariate Autoregressive modeling. *2009 IEEE/SP 15th Workshop on Statistical Signal Processing*, Cardiff, pp. 329-332. doi: 10.1109/SSP.2009.5278571.

[18] Luiz G. Hafemann, Robert Sabourin, Luiz S. Oliveira (2017), Learning features for offline handwritten signature verification using deep convolutional neural networks. *In Pattern Recognition*, Volume 70, Pages 163-176, ISSN 0031-3203, https://doi.org/10.1016/j.patcog.2017.05.012.

[19] Soleimani, A., Fouladi, K. and Araabi, B.N. (2017), UTSig: A Persian offline signature dataset. *in IET Biometrics*, vol. 6, no. 1, pp. 1-8, 1. doi: 10.1049/iet-bmt.2015.0058.

[20] Sepahvand, M., Abdali-Mohammadi, F. and Mardukhi, F. (2017), Evolutionary Metric-Learning-Based Recognition Algorithm for Online Isolated Persian/Arabic Characters, Reconstructed Using Inertial Pen Signals. *in IEEE Transactions on Cybernetics*, vol. 47, no. 9, pp. 2872-2884, Sept. doi: 10.1109/TCYB.2016.2633318.

[21] Naderi, N. and Nasersharif, B. (2017), Multiresolution convolutional neural network for robust speech recognition. *2017 Iranian Conference on Electrical Engineering (ICEE)*, Tehran, pp. 1459-1464. doi: 10.1109/IranianCEE.2017.7985272.

[22] Havaei, M., Davy, A., Warde-Farley, D., Biard, A., Courville, A., Bengio, Y., Pal, C., Jodoin, P.M., Larochelle, H. (2016), Brain tumor segmentation with Deep Neural Networks. *In Medical Image Analysis*, Volume 35, 2017, Pages 18-31, ISSN 1361-8415, https://doi.org/10.1016/j.media.2016.05.004.

[23] Su, T. W., Liu, J. Y. and Yang, Y. H. (2017), Weakly-supervised audio event detection using event-specific Gaussian filters and fully convolutional networks. *2017 IEEE International Conference on Acoustics, Speech and Signal Processing (ICASSP)*, New Orleans, LA, pp. 791-795. doi: 10.1109/ICASSP.2017.7952264.

[24] Dai, W., Dai, C., Qu, S., Li, J. and Das, S. (2017), Very deep convolutional neural networks for raw waveforms. *2017 IEEE International Conference on Acoustics, Speech and Signal Processing (ICASSP)*, New Orleans, LA, pp. 421-425. doi: 10.1109/ICASSP.2017.7952190.

[25] Long, J., Shelhamer, E., & Darrell, T. (2015). Fully convolutional networks for semantic segmentation. *In Proceedings of the IEEE Conference on Computer Vision and Pattern Recognition* (pp. 3431-3440).

[26] Huang, J. T., Li, J. and Gong, Y. (2015), An analysis of convolutional neural networks for speech recognition. *2015 IEEE International Conference on Acoustics, Speech and Signal Processing (ICASSP)*, South Brisbane, QLD, pp. 4989-4993.

[27] Sainath, T., Kingsbury, B., Saon, G., Soltau, H., Mohamed, A., Dahl, G. and Ramabhadran, B. (2015), Deep Convolutional Neural Networks for Large-scale Speech Tasks. *Neural Networks*, vol. 64, pp. 39-48.